# Interpretability and causal discovery of the machine learning models to predict the production of CBM wells after hydraulic fracturing

Chao Min[a, b, c, *], Guoquan Wen [a, b], Liangjie Gou[c], Xiaogang Li[c], Zhaozhong Yang[c]

[a] *School of Science, Southwest Petroleum University, Chengdu 610500, China*
[b] *Institute for Artificial Intelligence, Southwest Petroleum University, Chengdu 610500, China*
[c] *State Key Laboratory of Oil and Gas Reservoir Geology and Exploitation, Chengdu 610500, China*

## Abstract

Machine learning approaches are widely studied in the production prediction of CBM wells after hydraulic fracturing, but merely used in practice due to the low generalization ability and the lack of interpretability. A novel methodology is proposed in this article to discover the latent causality from observed data, which is aimed at finding an indirect way to interpret the machine learning results. Based on the theory of causal discovery, a causal graph is derived with explicit input, output, treatment and confounding variables. Then, SHAP is employed to analyze the influence of the factors on the production capability, which indirectly interprets the machine learning models. The proposed method can capture the underlying nonlinear relationship between the factors and the output, which remedies the limitation of the traditional machine learning routines based on the correlation analysis of factors. The experiment on the data of CBM shows that the detected relationship between the production and the geological/engineering factors by the presented method, is coincident with the actual physical mechanism. Meanwhile, compared with traditional methods, the interpretable machine learning models have better performance in forecasting production capability, averaging 20% improvement in accuracy.

## 1. Introduction

As an important supplement for the conventional energy sources, coalbed methane (CBM) is abundant in China. The total proved reserve of CBM in China is $6345\times10^8\text{m}^3$, among which the economical recoverable reserve is $2537\times10^8\text{m}$ [1]. The exploitation of CBM can not only help reduce the security risk during coal mining, but also regulate the structure of the energy supply of a country [2]. Due to the extremely low porosity and permeability of CBM reservoirs, a CBM well needs to be stimulated by hydraulic fracturing to obtain economical production [3-5], thus it is essential to optimize the fracturing engineering design by integrating the geological properties of the reservoir and the engineering conditions. Here it plays a key role to predict the production of the fractured CBM wells before the fracturing operation. There are many productivity-prediction models for CBM wells, including the seepage mechanism-based methods [8-11] and the data-based machine learning methods [6, 7].

---

[*] Corresponding author. School of Science, Southwest Petroleum University, Chengdu 610500, China. *E-mail address*: minchao@swpu.edu.cn (Chao Min).

In recent years, along with the development of big data techniques, machine learning has drawn much attention, such as multiple regression [8], random forest [9], support vector machine (SVM) and gradient-boosting [11] etc. However, their practical performance in field application are not satisfying enough. Good generalization performance of a machine learning model requires a sample data set with high quality and enough quantity, which is basically impossible to obtain in CBM development, as there are at most 300-400 wells in a CBM field. Thus, for the CBM wells with complicated geological/engineering conditions, the interpretability of the machine learning models is more attractive for predicting the production [12]. In artificial intelligence (AI), interpretability is one of the critical principles for trustworthy AI, stating that "for an AI system to be trustworthy, it must be able to behave in a certain way which is understandable to humans and provide an explanation to humans" [13].

In machine learning, causal discovery is defined as drawing the possible cause-and-effect relationships from observed data. Similar to interpretability, causal discovery also aims to explain the model for human [14]. A new perspective is proposed in this article focusing on the interpretability of the machine learning results to predict the output of CBM wells after hydraulic fracturing. Firstly, we present an improved causal discovery algorithm to investigate the truly relationship among production capability, geological/engineering factors. Subsequently, considering the input, output, treatment and confused variables deduced by causal discovery, we design an implicit causal predictive model for production capability. Moreover, to derive the explanation of the designed model, combining with the mechanism analysis, SHAP is employed to detect the influence of each factor impacting on production capability. Finally, to motivate the interpretability, causal variables are drawn to establish interpretable machine learning, including multiple regression, support vector machine, Multi-Layer Perceptron and random forest, for predicting-production model of CBM.

**Main Organization**
- The related works of the prediction methods of CBM production are discussed in Section 2, including the methods based on correlation, interpretability and causality, respectively;
- The improved causal discovery algorithm is demonstrated in Section 3;
- The application of the presented methods with the observed development data of a CBM reservoir is studied in Section 4, including the improved causal discovery for global and local causality, SHAP for interpretability, and an interpretable machine learning under causality for predicting-production;
- Section 5 is the conclusion of this article.

**Main Contribution**
- To our knowledge, this is the first presence using causal science to discuss the interpretability of the machine learning models in CBM development;
- Not like the correlation analysis, we focus on finding the causality between the production capability and the geological/engineering factors;
- A novel causal discovery algorithm is provided in this article and we design an implicit causal predicting-production to investigate the underlying causality from observed data, the result of which basically coincides with mechanism and human cognition;

## 2. Related Works

There are many researches focus on forecasting production capability of CBM wells and finding its controlling factors. The methods can be classified into 3 categories, the methods based on correlation [16-19], methods with interpretability [8-11, 24-16], and with causality [32-37].

**The Methods based on Correlation.** Correlation in this article refer to finding the quantitative indicators that can reflects the mutual relation between the varying factors [15]. Liu H et al. [16] apply information entropy method to evaluate the correlation between the geologic factors and the CBM recovery from coal mine, and establish the forecasting model based on fuzzy theory. In [17], bivariate correlation analysis was applied to determine the interrelationships between temporal and spatial factors, including burial depth, thickness of coal, gas content, porosity/permeability, effect of fracturing, structural setting, and hydrogeological conditions. And then Lv Y et al. applied the gray system theory to predict the CBM well productivity. Taking in count the influence of geological factors regarding coalbed methane, Wang G et al. [18] obtain the correlation of the quantitative and qualitative index, and then evaluated the production through the fuzzy theory. To deal with the simulation of data with more complicated characteristics, the quantity $k^2d$ and $r$-order was applied to analysis the correlation between variables, and then Zeng B et al. [19] constructed a forecasting model for CBM production by grey system.

**The Methods with Interpretability.** Interpretability is the key of a responsible and open data-science, across multiple industry areas and scientific disciplines [14]. Different from traditional machine learning, interpretable machine learning can not only provide the output of the model, but also explain the feature-attribution mechanism between the input and output of model. In the existing interpretable models [20-22], decision tree, rules, linear models [23] are interpretable for humans. Random forests, boosted trees are the typical decision-tree ensemble methods, Support Vector Machine (SVM) is a method searching for hyperplanes with decision boundary based on a set of established rules. There are some interpretable researches for predicting production in CBM. To avoid the multicollinearity caused by the improper selection of the independent variables, Pearson correlation was applied to optimize Random Forest in [9], which can prompt the controlling factors of CBM production. Utilizing all features, including geological/engineering factors, Erofeev A S et al. [11] applied gradient-boosting to predict the production. Guo Z et al. [10] designed a regressive SVM model with all features to approximate the reasonable production from the reservoir-simulation model, guiding for the field-scale. Based on the selected controlling factors that affect the production of shale gas wells, multiple linear regression method was established for predicting production of shale gas wells in [8]. Neural Network is a method can simulate complex non-linear relationship by multiple hidden layers and non-linear model is an interpretable function simulating the outputs by the nonlinear combination of the parameters. Multilayer perceptron (MLP), back propagation (BP) neural network and long short-term memory (LSTM) yields to the interpretable neural network. Based on the data-driven theory, MLP [24, 25] was developed to understand the static and dynamic features in the fractured reservoir, then to predict the production. Due to the heterogeneity of coalbed, the uniqueness of CBM production process, the complexity of data's features, Lü Y M et al. [26] utilize the BP neural network to deal with the matching the past gas production and predicting the futural production performance. Under the influence of the geological/engineering factors in CBM, the data-driven production forecasting model based on the LSTM neural network are designed [27, 28] for CBM, which have more powerful adaptability and accuracy for both regular and irregular production behavior. In nonlinear machine learning models, e.g., deep neural networks, it is difficult to directly perceive the interpretability of model.

Therefore, model-agnostic methods are always utilized to explain the model. SHAP [29] is a popular feature-attribution mechanism for interpretable machine learning, which applies game-theoretic notions to measure the attribution of each factor on output. SHAP has attracted generous attention of academia and industry recent year [13].

**The Methods with Causality.** The causality among geological/engineering factors and production capability is extremely useful for CBM development. However, it is difficult to discover causality purely from observed data, and there is no relevant research in CBM. Numerical simulation is a typical data-driven method under interpretability [12, 30, 31]. When numerical simulation can effectively approximate the existing mechanism, it has latent causality. Vishal V et al. [32] use the finite difference based numerical simulations to conduct the assessment of the factors effecting the long-term CBM recovery in Jharia coalfield, including matrix carbon dioxide concentration, fracture gas saturation, matrix methane concentration and total water production. To enhance the recovery of coalbed methane, a finite difference numerical model [33] was proposed to simulate the data from one western China basin, which can increase the production of coalbed methane. Nie B et al. [34] designed a model consisting of momentum conservation equation and energy conservation equation, and then utilized the numerical simulation to obtain its numerical solution. The influence of the injection parameters for the development of coalbed methane reservoirs can be analyzed. Similarly, many researches [35-37] design the production predicting model for coalbed methane based numerical simulation methods. These methods have interpretability and causality in some aspects, but require a large number of model parameters and high accuracy of input data. Moreover, the limitation of numerical simulation is that it requires the observed data to satisfy the corresponding theory and assumptions, such as in fluid dynamics. The whole investigation is presented in Table 1.

**Table 1.** The methods with different explanation.

| Model | Correlation | Interpretability | Causality |
|---|---|---|---|
| Correlation Analysis [17] | ✓ | | |
| Grey System [19] | ✓ | | |
| Fuzzy system [16, 18] | ✓ | | |
| SVM [10] | ✓ | ✓ | |
| Gradient-Boosting [11] | ✓ | ✓ | |
| Random Forest [9] | ✓ | ✓ | |
| Multiple Regression [8] | ✓ | ✓ | |
| MLP [24, 25] | ✓ | ✓ | |
| BP Neural Network [26] | ✓ | ✓ | |
| LSTM [27, 28] | ✓ | ✓ | |
| Numerical Simulation [32-37] | ✓ | ✓ | ✓ |
| Our method | ✓ | ✓ | ✓ |

*Note.* ✓ denotes the method take the corresponding property, including correlation, interpretability and causality.

Causal inference is an emerging discipline aiming to explore causality in machine learning [38, 39]. Causal discovery has drawn much attention in machine learning, statistics, medicine, economics and climatology [40-43] and has shown better interpretability and robustness. To our knowledge, none of the existing methods apply causality to explore the main controlling factors and forecast production capacity in CBM.

# 3. The Improved Iterative Causal Discovery (IICD) Algorithm

In this section, we introduce some relevant background on causality theory, including basic graphs, definitions and conditions[1], and then we will propose our method, improved iterative causal discovery (IICD).

## 3.1 The Definition and Condition in Causal Discovery

**Causal Graph.** The goal of causal discovery is to understand the causality among the variables, and the causality is mainly reflected by causal graph. There are three basic graphs for the causal discovery in Fig. 1, and its specific application has represented in Algorithm 1.

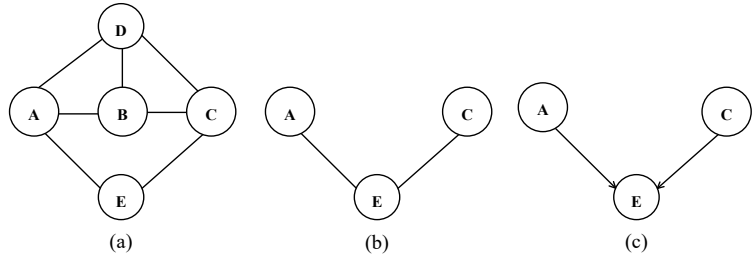

**Fig. 1**. (a): An undirected graph; (b): one of skeleton of (a), in which the selected node is *E* and its adjacent nodes are *A* and *C*; (c): the v-structure for the node *E*, with respect to *A* and *C* ($A \rightarrow E \leftarrow C$).

In causal discovery, the causality among the variables generally explores several different graphical representations, including partial ancestral graph (PAG), directed acyclic graph (DAG), and maximal ancestral graph (MAG), as shown in Fig. 2.

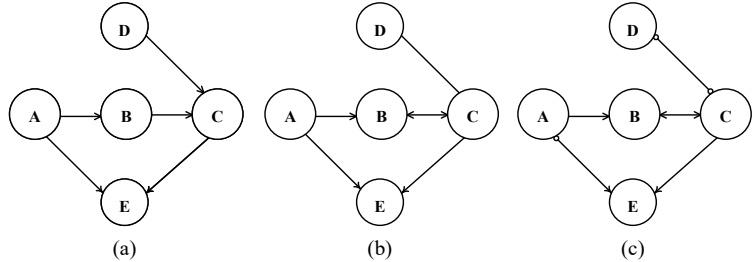

**Fig. 2**. Three typical causal graphs. (a): DAG, there is only one type for edge $\longrightarrow$; (b): MAG, there are only three types for edge, including $\longrightarrow, \longleftrightarrow$ and —(c): PAG, there are only four types for edge, including $\longrightarrow, \longleftrightarrow$, o— and o—o.

Totally, there are five possible causalities very pair of nodes $x$ and $y$ in an $edge(x, y)$:
- $\leftarrow$ or $\rightarrow$ represents that a node is a cause of another node;
- — represents that the nodes are neighbors;
- $\longleftrightarrow$ represents that there is a latent common cause for nodes;
- o— represents that the node is not the cause of the node close to 'o';
- o—o represents that there is no D-separation between nodes.

Causal discovery is an Algorithm based on Bayesian theory. Before presenting the Algorithm in detail, some definitions and conditions need to be provided.

**Definition 1 (D-Separation):** *A path P is blocked by a set of nodes **Z** if and only if*
*1. P contains a chain of nodes $A \rightarrow B \rightarrow C$ or a fork $A \leftarrow B \rightarrow C$ such that the middle node B is in **Z** (i.e., B is conditioned on), or*

---
[1] More details of graphs, definitions and conditions with respect to causal discovery, please refer to references [14]

2. P contains a collider $A \rightarrow B \leftarrow C$ such that the collision node B is not in **Z**, and no descendant of B is in **Z**.

*Then, we say A and B are D-separated by **Z** and denote the D-Separation on a DAG $\mathcal{G}$ respect to the A, B and C as $\perp\!\!\!\perp_{\mathcal{G}}$.*

Some conditions with respect to the causal discovery algorithm should be satisfied, including causal sufficiency, causal Markov condition, faithfulness and independence tests, which are as follows.

**Condition 1 (Causal Sufficiency).** *For every pair of the observed variables in a given dataset, all common causes can be observed in the dataset.*

**Condition 2 (Causal Markov Condition).** *A directed acyclic graph $\mathcal{G}$ over vertexes set **V** and a probability distribution $\mathcal{P}$ satisfy the causal Markov condition if and only if for every $x \in \mathbf{V}$, $x \perp\!\!\!\perp_{\mathcal{G}} \mathbf{V} \setminus (\mathrm{De}(x) \cup \mathrm{Pa}(x))$, where $\mathrm{De}(x)$ and $\mathrm{Pa}(x)$ denote the descent and parent node of x, respectively.*

**Condition 3 (Faithfulness).** *Let DAG $\mathcal{G} = (\mathbf{V}, \mathbf{E})$ be a causal graph and $\mathcal{P}$ a probability distribution generated by $\mathcal{G}$. $\langle \mathcal{G}, \mathcal{P} \rangle$ satisfies the Faithfulness Condition if and only if e $x \perp\!\!\!\perp_{\mathcal{G}} y|\mathbf{V} \Leftarrow x \perp\!\!\!\perp_{\mathcal{P}} y|\mathbf{V}$, where $x, y \in \mathbf{V}$ and **E** denote the edges set.*

**Condition 4 (Conditional Independence Tests).** *Conditional independence tests are satisfied when correct independence test is applied to corresponding variable and if the test has a reliable result.*

If all the above conditions are satisfied, a causal discovery algorithm can be designed. The typical algorithm can be divided into three steps: constructing an initial causal graph, removing some unreasonable edge in graph, and orienting the edge in graph. The details are listed as follows in Algorithm 1.

| **Algorithm 1.** A Basic Causal Discovery Algorithm |
|---|
| **Input:** $D^{n \times d}$ (A conditional independence Data with size $n \times d$) |
| **Output:** $\mathcal{G}$ (A completed DAG with Markov equivalence class) |
| **Step 1:** There are two different approaches:<br>• *building an **undirected complete graph** with all variables; then iterating with an arbitrary node in graph. In each iteration, removing or remaining each edge through conditional independence tests, e.g., Fig. 1(a);*<br>• *searching the **local skeleton** (e.g., Fig. 1(b)), for the selected variables; finding their adjacent nodes to form a local skeleton. Next, aggregating all local skeletons into a global skeleton, e.g., Fig. 1(a);*<br>**Step 2:** *Finding the connections that have **v-structure** with the result in Step 1(e.g., Fig. 1(c)). And if there are two connections of type $A \rightarrow E$ and $C \rightarrow E$, then removing the edge in A and C, which is done by **D-separation**.*<br>**Step 3:** *Orienting the remained undirected-edges in terms of established rules and observed data, and forming a **causal graph** in Fig. 2, finally.* |

## 3.2 The Framework of IICD

A causal graph is a typical sparse graph as the causality between factors is unidirectional and scarce. Unlike the principles in iterative causal discovery (ICD) [44], causal discovery recovering sparse graph is not regarded as an NP-hard problem [45]. During the recovery, redundant information not involving sparse graph will disturb the determination of causal relationship for each pair of the nodes. To deal with this problem, an improved method is proposed to recovery the causality for a sparse graph, named Improved Iterative Causal Discovery (IICD). In each iteration, IICD will gradually refine the causality of each pair of the nodes to mitigate the noise disturbance leaded by redundant information.

The goal of IICD is to discover a truly underlying DAG in the presence of a series of variables, which can be represented by an equivalent causal graph PAG $\mathcal{G}$ in terms of Faithfulness. Our method is endued to detect and optimize causal absence leaded by sparse structure, in which not only the adjoint nodes of $y$ but also the adjoint nodes of $x$ are considered with respect to a $edge(x, y)$, see line 4 of Algorithm 2.

IICD Algorithm starts with $r = 0$ and updates with $r = r + 1$ in each iteration. Same as the first Step of Algorithm 1, IICD starts with an undirected complete graph and the graph is gradually refined by the correct conditional independence (CI) tests. This loop is in lines 2-16, until a global skeleton is obtained. Similar to the Step 2 in Algorithm 1, IICD searches all possible latent v-structure to maintain or remove the edge in v-structure by combining $D$-separation with PDS-path $\Pi_y(x, z)$ (shown in definition 3). Finally, IICD orient the edge in $\mathcal{G}$ with five causal representations until Step 1 and Step 2 end up with $r = d - 2$.

**Algorithm 2. Iterative Causal Discovery (ICD)**

**Input:** $D^{n \times d}$ (A conditional independence Data with size $n \times d$)
**Output:** $\mathcal{G}$ (A completed PAG)

1: **Initialize:** $r \longleftarrow 0$, $\mathcal{G} \longleftarrow$ an undirected complete graph with 'o' edge-mark
2: **while** $r \leq d - 2$ **do** ▷ *Step 1 in Algorithm 1 start*
3:     **for** $edge(x, y)$ in $\mathcal{G}$ **do**
4:         get subset $\mathbf{Z}_x \subseteq \text{Adj}(x) \setminus y$ and $\mathbf{Z}_y \subseteq \text{Adj}(y) \setminus x$, where $|\mathbf{Z}_x| = |\mathbf{Z}_y| = r$
5:         create $\Pi_y(x, z_x), z_x \in \mathbf{Z}_x$; $\Pi_x(y, z_y), z_y \in \mathbf{Z}_y$ ▷ *Step 2 in Algorithm 1 start*
6:         $\mathbf{Z}_i \longleftarrow (\Pi_y(x, z_x) \cup \Pi_x(y, z_y)) \setminus \{x, y\}$
7:         order $\{\mathbf{Z}_i\}_{i=1}^l$ by $d(\mathbf{Z}_i) = \frac{1}{|\mathbf{Z}_i|} \sum_{z \in \mathbf{Z}_i} \min(|\Pi_y(x, z)|, |\Pi_x(y, z)|)$
8:         **for** $\mathbf{Z}_i$ in $\{\mathbf{Z}_i\}_{i=1}^l$ **do**
9:             **if** $x \perp\!\!\!\perp_\mathcal{G} y | \mathbf{Z}_i$ **then**
10:                 $\mathcal{G} \longleftarrow$ remove $edge(x, y)$ from $\mathcal{G}$
11:                 $Sepset(x, y) = Sepset(y, x) \longleftarrow \mathbf{Z}_i$
12:                 $\mathbf{Z}_i \longleftarrow Dsepset(x, y, \mathcal{G})$
13:             **end if**
14:         **end for** ▷ *Step 2 in Algorithm 1 end*
15:         $r \longleftarrow r + 1$
16:     **end for** ▷ *Step 3 in Algorithm 1 start*
17:     $\mathcal{G} \longleftarrow$ orient the edges with five causalities ▷ *Step 3 in Algorithm 1 end*
18: **return** $\mathcal{G}$ ▷ *Step 3 in Algorithm 1 end*

**Note:** $\text{Adj}(x) \setminus y$ denote the adjoint node of $x$, except $y$; $|\cdot|$ denote the number of variables; $d(\mathbf{Z}_i)$ denote the ordering criteria for node set $\mathbf{Z}_i$; $Sepset(x, y)$ denote the separation set for nodes $x$ and $y$; $Dsepset(x, y, \mathcal{G})$ denote the D-separation set with respect to $\mathcal{G}$.

**Definition 3** (PDS-path). When causal Markov condition is satisfied, a possible-D-Sep-path (PDS-path) from node $x$ to node $z$, with respect to $y$ in a DAG, denoted by $\Pi_y(x, z)$, is a path $\langle x, \cdots, z \rangle$ such that $y$ is not on the path and for every sub-path $\langle u, v, w \rangle$ of $\Pi_y(x, z)$, $v$ is a collider or the node set $\{u, v, w\}$ forms a triangle.

To be more specific, we provide a tiny example of IICD with iteration from $r = 0$ to $r = 1$ shown in Fig. 3. When $r = 0$, namely, the first iteration of IICD, the initial graph is an undirected complete graph with 'o' edge-mark. When each pair of the nodes in an edge of Fig. 3 satisfy the conditional independence (CI) tests, the initial graph will be converted into an initial causal graph, that is, Fig. 3 (a). In the second iteration with $r = 1$, taking the $edge(D, E)$ as an initial condition for Fig. 3 (b), there are only the nodes that are adjacent to node $D$ or $E$ composed with a conditioning set, including $A$, $B$ and $C$. By utilizing the PDS-path and D-separation, v-structure in graph is found and the corresponding edges are removed, including the $edge(D, C)$ and $edge(A, B)$, compared with Fig. 3 (a). Shown in Fig. 3 (c), IICD orients the edges in Fig. 3 (b) with five possible causalities.

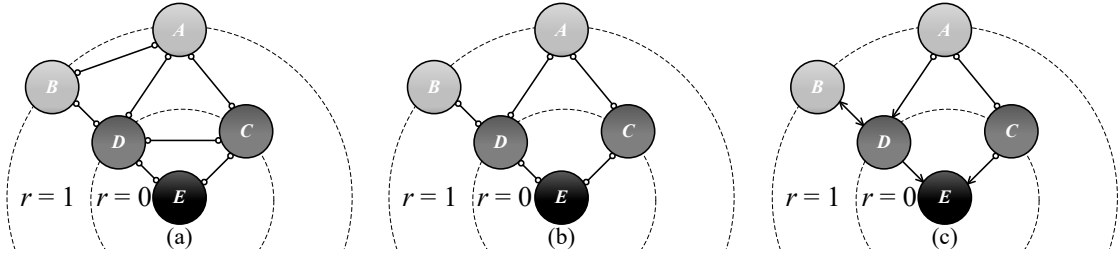

**Fig. 3**. A tiny example of constructing a causal graph built by the IICD with $r = 0 \Longrightarrow r = 1$. (a) Searching and Building the skeletons for $edge(D, E)$ by CI tests; (b) Removing or remaining the edges in the (a) according to a series of rules and conditions in IICD. Compared with (a), there are four edges have been remained, including $edge(D, E), (E, C), (B, D)$ and $(A, D)$. Moreover, the $edge\ (B, A)$ and $(D, C)$ have been removed.); (c) A PAG built by orienting the remained edges in (b) according to rules and conditions of IICD, e.g., $B$ o—o $D \Rightarrow B \leftrightarrow D$ and $C$ o—o $E \Rightarrow C$ o$\rightarrow E$.

## 4. The Causality and Interpretability in Observed Data of CBM

In this article, we collect the data from 408 CBM wells after hydraulic fracturing. The data of each well consists of geological/engineering factors and production capability. To investigate the truly relationship between factors and gas production over the observed data, IICD is applied to investigate the causality in CBM and the result is presented as a causal graph (in subsection 4.1 and 4.3). Then SHAP is utilized to measure each factor's contribution on output with respect to the designed causal prediction production model, means interpretability (in subsection 4.2). Finally, we apply the machine learning method to predict the production (in subsection 4.4) based on the owned causality. The more details of application and results are represented in following subsection.

### 4.1 The Global Causality in Observed Data of CBM

The underlying causality existing in factors of CBM is obviously unidirectional and scarce, which means its causal graph is a typical sparse graph. By using the IICD, we obtain its global causality between geological/engineering factors and production capability. As shown in Fig. 4, rather than human-made intervention, a potential causality of observed data in CBM has been recovered, which is a typical PAG. With limited data, PAG truly reflects the underlying causality among the geological/engineering factors and the gas production in CBM wells after hydraulic fracturing.

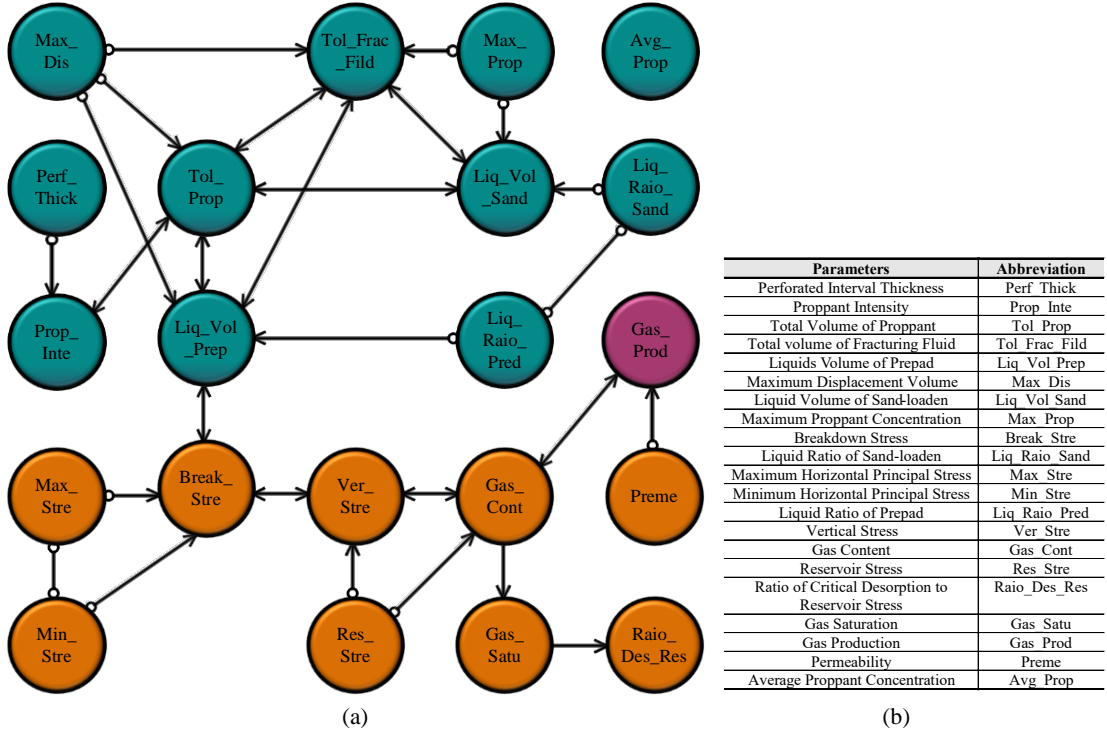

**Fig. 4.** The global causal graph PAG of CBM and its symbol description. (a): recovered PAG over the observed data of CBM wells, in which geological/engineering factors and production capability are marked by dark-yellow, dark-green and dark-red, respectively; (b) the abbreviation of factors and gas production with respect to fig. 4(a).

In Fig. 4, it apparently separates the geological and engineering factors into two relatively independent parts in terms of dark-yellow and dark-green circles. However, not in isolation, geological and engineering factors are associated by $Liq\_Prep \longleftrightarrow Break\_Stre$, namely, the $edge\ (Liquids\ Volume\ of\ Prepad, Breakdown\ Stress)$. To promote the gas production of CBM wells, causality represented in PAG of Fig. 4 demonstrate that we must consider whether the rocks of the reservoir can be broken down or not by pumping the fracturing liquid. Moreover, we can know that the gas production is mainly benefited from the geological factors, which is also innate resource advantages for a CBM. The engineering factors affect gas production through Breakdown Stress.

Causality focuses on cause and effect in the observed data of CBM, and cannot provide the factor-contribution mechanism. To evaluate the magnitude of causality, we applied SHAP to investigate the contribution of each factor impacting on model output, that is, interpretability. In the following, we will illustrate the rationality of the interpretable results by combining global and local causality with the theory in fracturing engineering.

### 4.2 The SHAP under Causality

To explain the causal graph of Fig. 4 in terms of interpretability, we firstly design a nonlinear causal production-predicting model for CBM based on the causality in Fig. 4. The model is formulated as,

$$\begin{cases} T = \psi_1(W) + \epsilon_1 & (5.a) \\ Y = \psi_2(\gamma(X) \cdot T + \beta(X, W)) + \epsilon_2 & (5.b) \end{cases} \quad (1)$$

where $\psi_1$ and $\psi_2$ are the fitting model for $T$ and $Y$, respectively; $\epsilon_1$ and $\epsilon_2$ are the bases; $\gamma(\cdot)$ and $\beta(\cdot)$ denote the implicit function for variables $X, W$; $W$ denotes one of causal set for $T$; $X$ and $W$ are both one of causal set for $Y$; $T$ is not only an intervention variable but also a cause for $Y$.

Obviously, there is a main causal chain in Fig. 4, see Fig. 5.

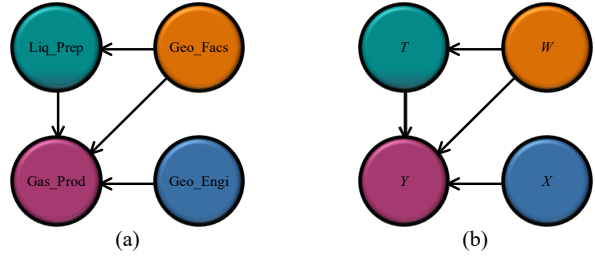

**Fig. 5**. The main causal chain with respect to Fig. 4. (a) The streamlined causal structure of Fig. 4, where Geo_Facs denotes the geological factors marked by dark-yellow and Geo_Engi denotes the factors consisted of geological and engineering factors, except *Liquid Volume of Prepad*, marked by dark-blue; (b) the specific causal structure with respect to Eq. (1) and Fig. 5(a).

Furthermore, we can find that the geological factors are the cause of *Gas Production* and engineering factors; engineering factors are also the cause of *Gas Production* under the intervention variable *Liquid Volume of Prepad* ($Liq\_Prep$). Moreover, *Liquid Volume of Prepad* affects *Gas Production* through *Breakdown Stress* ($Break\_Stre$). Then, by using the causal chain in Fig. 5(a) and its properties, we propose a causal production-predicting model combing the causal explainer, that is, the $\psi_i, i = 1,2$ in Eq. (1). The proposed model should be subjected to the following principles:
- $T$ is the *Liquid Volume of Prepad*, namely treatment variable;
- $W$ consists of all variables in geological factors, namely confusing variables;
- $X$ consists of all variables in geological and engineering factors, except $T$, namely input variables;
- $Y$ is the gas production, namely output variables;
- $\psi_i, i = 1,2$ denote the causal explainer for Treatment $T$ and Output $Y$, in which the explainer will be replaced by random forest, MLP, SVR and linear regression.

SHAP can explain the feature-attribution mechanism of the model by computing the contribution of each feature on output, which is represented as an additive feature-attribution linear model. It specifies the explanation as:
$$f(x') = \phi_0 + \sum_{i=1}^{n} \phi_i x'_i \quad (2)$$
where $f$ denote the explaining model, $x' \in \{0,1\}^n$ denote the feature can be observed or not (1 or 0), $n$ is the number of the features, $\phi_0$ is a constant for explanation, and $\phi_i \in \mathcal{R}$ is the Shapley value of the $i$-th feature.

Shapley values is a method from coalitional game theory, which provided a standard of how to fairly distribute the *payout* among the features to determine the precise output. The Shapley value $\phi_i$ of the $i$-th feature is defined as
$$\phi_i = \sum_{S \subseteq \{x_1, \cdots, x_n\} \setminus \{x_i\}} \frac{|S|!\,(n-|S|-1)!}{n!} (f_x(S \cup \{x_i\}) - f_x(S)) \quad (3)$$
where $\{x_i\}_{i=1}^n$ denotes the features, $f_x(S)$ is the prediction for the subset $S$.

To investigate the contribution of each factor including geological and engineering factors, on *Gas Production*, SHAP is utilized to give its explanation in terms of the causal model in Eq. (1). The SHAP result is shown in Fig. 6, which mainly reflects the factor importance with factor effects, that is, factor effects on production are sorted in descending order by factor importance. Moreover, each point in Fig. 6 denotes a Shapley value for a factor of a CBM well. The color blue, green and red represent the low, medium and high value of corresponding factor. And its $x$-axis represents the Shapley value. Intuitively, factor importance is a kind of interpretability, which is understandable to humans.

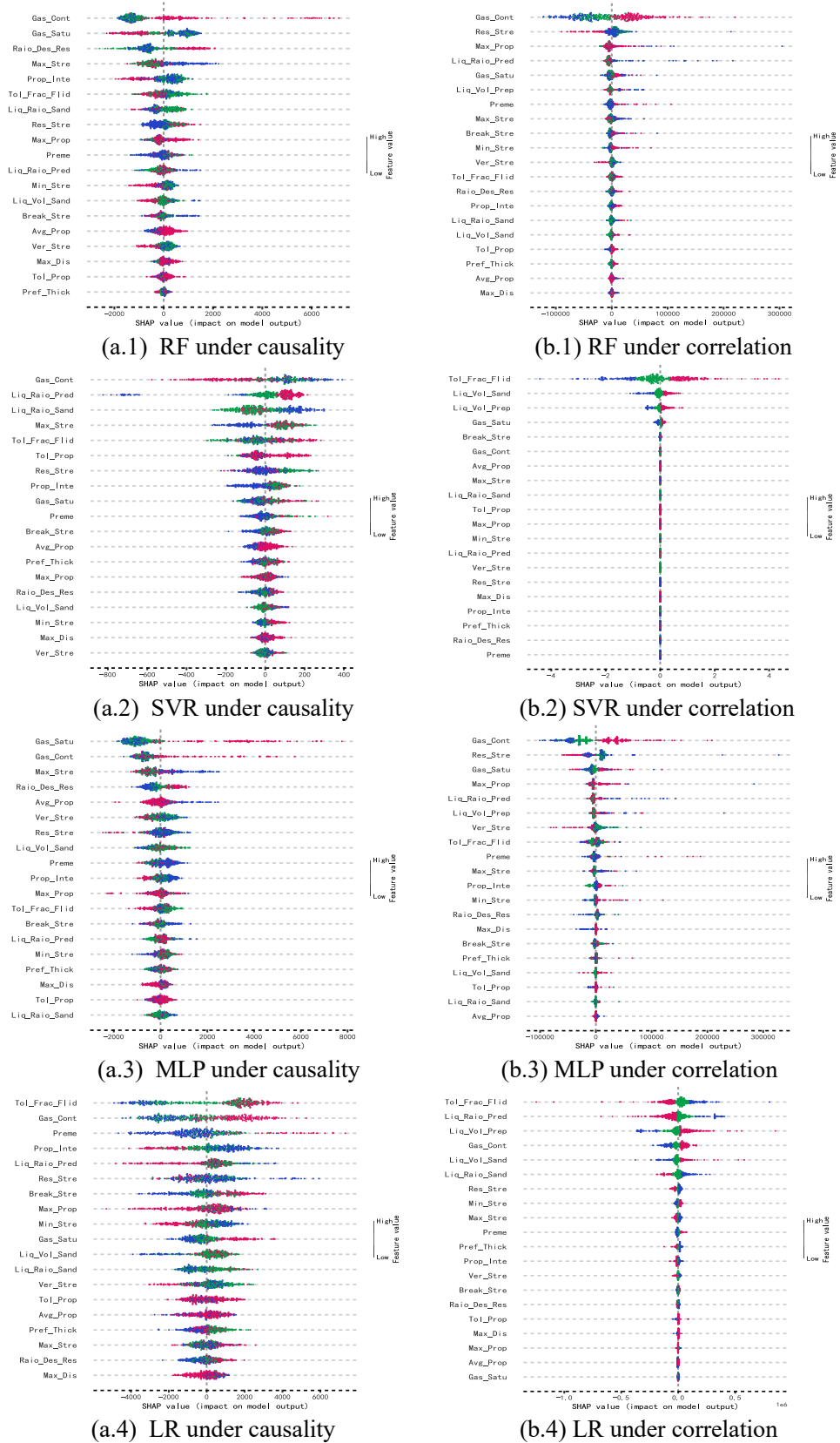

(a.1) RF under causality    (b.1) RF under correlation

(a.2) SVR under causality    (b.2) SVR under correlation

(a.3) MLP under causality    (b.3) MLP under correlation

(a.4) LR under causality    (b.4) LR under correlation learning. Fig. 6(a.∗) and 6(b.∗) denote the SHAP value under causality and correlation, respectively. Fig. 6(∗.1-4) denotes the SHAP value with machine learning method, that is, **random forest**, **SVR**, **MLP** and **linear regression**, respectively.

**Fig. 6.** The SHAP value impacting gas production under causality and correlation with machine

As shown in subfigure of Fig. 6, the factors affecting production are sorted in descending order by its contribution on *Gas Production*, e.g., the contribution of *Gas Content* on *Gas Production* is more than *Gas Saturation*, that is, the interpretability of *Gas Content* on *Gas Production* is more reliable than *Gas Saturation* in Fig. 6(a.1). Furthermore, SHAP under causality provide more specific trend of interpretability for each factor impacting on *Gas Production*. With causality, there are four patterns of trends between the nodes in the casual graph. In Fig. 6 (a.1), we can see that higher value of *Gas Content* indicates higher value of *Gas Production*, this is denoted by ***O***; in Fig. 6 (a.2), higher value of *Maximum Horizontal principal Stress* indicates lower value of *Gas Production*, this is denoted by ***N***; in Fig. 6 (a.3), moderate value of *Gas Saturation* indicates higher value of *Gas Production*, this is denoted by ***M***; in Fig. 6 (a.4), *Average Proppant Concentration* will confuse the value of *Gas Production*, this is denoted by ***C***. The complete result of all factors contribution has been represented in Table 2.

Table 2. The interpretability determined by SHAP with causal explainer

| Geological/Engineering Factors | LR | SVR | MLP | RF |
|---|---|---|---|---|
| Perforated Interval Thickness | *O* | *O* | *O* | *O* |
| Proppant Intensity | *M* | *O* | *N* | *M* |
| Total Volume of Proppant | *C* | *O* | *C* | *O* |
| Total Volume of Fracturing Fluid | *O* | *O* | *M* | *M* |
| Liquid Volume of Prepad | *C* | *C* | *O* | *O* |
| Maximum Displacement Volume | *C* | *O* | *C* | *O* |
| Liquid Volume of Sand-Loaden | *O* | *N* | *M* | *M* |
| Maximum Proppant Concentration | *C* | *C* | *C* | *O* |
| Breakdown Stress | *O* | *O* | *N* | *O* |
| Liquid Ratio of Sand-Loaden | *O* | *N* | *C* | *M* |
| Maximum Horizontal Principal Stress | *C* | *O* | *N* | *N* |
| Minimum Horizontal Principal Stress | *M* | *O* | *O* | *M* |
| Liquid Ratio of Prepad | *C* | *O* | *N* | *M* |
| Vertical Stress | *M* | *O* | *N* | *M* |
| Gas Content | *O* | *N* | *O* | *O* |
| Reservoir Stress | *C* | *M* | *N* | *O* |
| Ratio Of Critical Desorption to Reservoir Stress | *M* | *O* | *O* | *O* |
| Gas Saturation | *O* | *O* | *O* | *M* |
| Permeability | *O* | *O* | *M* | *M* |
| Average Proppant Concentration | *C* | *O* | *C* | *C* |

*Note. **O** and **N** denote the corresponding factor take the optimistic and negative interpretability for Gas Production; **M** denotes the corresponding factor take the optimistic interpretability when there is appropriate value; **C** denotes the corresponding factor take the confusing interpretability.*

The complete interpretability of our causal predicting-production model is demonstrated in detail in Fig. 6(a.1-a.4) and Table 2. To prove the superiority of our method in revealing the factor-mechanism of CBM, the SHAP under correlation is compared with the SHAP under causality, the results of which is shown in Fig. 6 (b.1-b.4). From Fig. 6 (b.1-b.4), we can see that the SHAP under causality leads to more clustered Shapley values for each factor, which reflects less heterogeneity, especially for the factors with lower contribution. Thus, we can say that the presented method can improve the interpretability of machine learning models.

The interpretability in Table 2 with causal variables is almost consistent of the human cognition to the knowledge in fracturing engineering. Additionally, from the comparison in Fig. 6, it reports that the random forest under causality provides the most reasonable interpretability.

## 4.3 The Local Causal Discovery

In subsection 4.1 and 4.2, we have discussed the rationality of our method in a global view. In this subsection, we will further investigate the causality from a local perspective. Benefiting from the global causal graph in subsection 4.1, we see that *Liquid Volume of Prepad* and *Breakdown Stress* are the critical factors for the *Gas Production*. Subsequently, we extract a shortest critical causal path that involves at least one factor in geological/engineering scope with *Gas Production*, as shown in Fig. 7.

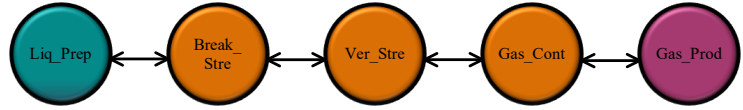

**Fig. 7.** The shortest critical causal path with critical factors for *Gas Production*. *Liquid Volume of Prepad* yields engineering factors; *Breakdown Stress*, *Vertical Stress* and *Gas Content* yields geological factors.

**Table 3.** Nomenclature of factors[2]

| Factors | Symbols | Factors | Symbols |
| --- | --- | --- | --- |
| The Breakdown Stress | $p_F$ | The Overlying Formation pressure | $S_V$ |
| The Pore Pressure | $p_s$ | The Pore Pressure Coefficient | $A_{pe}$ |
| The Tensile Strength | $\sigma_t$ | The Initial Principal Stress | $S_{hi}$ |
| The Poisson's Ratio | $v$ | The Total Pressure Loss | $\Delta p_T$ |
| The Fracture Length | $L$ | The Bottom Hole Flow Pressure | $p_{wf}$ |
| The Fracture Height | $H$ | The Gas Layer Thickness | $h$ |
| The Rock Cut Modulus | $G$ | The Formation Volume Factor of Gas | $B_g$ |
| The Fracture Width | $W$ | The Gas Viscosity Coefficient | $\mu_g$ |
| The Proppant | $u_p$ | The Reservoir Temperature | $T$ |
| The Coefficient | $\alpha$ | The Initial Reservoir Temperature | $T_0$ |
| The Gas Production | $q_g$ | The Initial Reservoir Stress | $p_0$ |
| The Permeability | $K$ | The Quantity of Gas Adsorption | $V_p$ |
| The Reservoir Stress | $p_e$ | The Gas Compressibility Factor | $Z$ |
| The Supply Radius | $r_e$ | Total volume of Fracturing Fluid during Engineering | $Q_f$ |
| The Well Radius | $r_w$ | The Liquid Volume of Prepad | $Q_p$ |
| The Skin Coefficient | $S$ | The Liquid Volume of Sand-loaden | $Q_s$ |
| The Porosity | $\phi$ | The Liquid Volume of Displacing | $Q_r$ |
| The Gas Saturation | $S_g$ | Maximum Horizontal Principal Stress | $\sigma_y$ |
| The Gas Content | $S_c$ | Minimum Horizontal Principal Stress | $\sigma_x$ |

***Proof for the edge*** (*$Liquids\ Production\ of\ Prepad, Breakdown\ Stress$*). In Fig. 7, there is an edge $Liq\_Prep \longleftrightarrow Break\_Stre$, which means there exist latent common causes between *Liquid Volume of Prepad* and *Breakdown Stress*. And we will prove the conclusion from the mechanism in Fracturing Engineering.

The equation of *Breakdown Stress* $p_F$ is defined as,

$$p_F = \frac{2\left(\frac{v}{1-v}\right)S_V + 2S_{hi} + A_{pe}p_s + \sigma_T}{2 - A_{pe}} \Longrightarrow p_F = f_1(S_V, v, S_{hi}, A_{pe}, p_s, \sigma_T) \quad (4)$$

And the equation of *Fracture Width* $W(x,t)$ is defined as,

---
[2] Nomenclature of each factor in Eqs. (4)-(14)

$$\begin{cases} W(x,t) = W(0,t) \left( \frac{x}{L} \sin^{-1} \frac{x}{L} + \sqrt{1 + \left(\frac{x}{L}\right)^2} - \frac{\pi}{2} \frac{x}{L} \right)^{\frac{1}{4}} \\ W(0,t) = \frac{(1-v)\Delta p_T H}{G} \end{cases} \quad (5)$$

$$\Longrightarrow W(x,t) = f_2(x,t,L,v,\Delta p_T, H, G) \quad (6)$$

One of the goals of employing Prepad is to make the fracture width wide enough so that the proppant can enter the artificial fractures [46], that is, the $W(x,t)$ and the Proppant $u_p$ are the causes of Prepad $Q_p$, then we have

$$Q_p = f_3(W, u_p) \Longrightarrow r_p = f_3(u_p, x, t, L, v, \Delta p_T, H, G) \quad (7)$$

According to the Eq. (4) and (7), there exists a common cause the Poisson's ratio $v$ for the $p_F$ and $Q_p$, then it approves the causal relation $Liq\_Prep \leftrightarrow Break\_Stre$ shown in Fig. 7.

**Proof for the edge ($Gas\ Content, Gas\ Production$).** In Fig. 7, there is also an edge $Gas\_Cont \leftrightarrow Gas\_Prod$, which means there exists latent common causes $v$ between the *Gas Content* and *Gas Production*. And we will give the proving process from the theory in Fracturing Engineering.

The equation of *Gas Production* $q_g$ can be computed by the following formula,

$$q_g = \frac{\alpha K h}{B_g \mu_g \ln \left(\frac{r_e}{r_w} + S\right)} (p_e - p_{wf}) \Longrightarrow q_g = g_1(K, h, p_e, p_{wf}, B_g, \mu_g, r_e, r_w, S) \quad (8)$$

The equation of *Gas Content* $S_c$ is

$$S_c = \frac{\phi S_g p_e T_0}{p_0 T Z} + V_p \Longrightarrow S_c = g_2(\phi, S_g, p_e, T, Z, V_p) \quad (9)$$

According to the Eq. (8) and (9), there exists a common cause *Reservoir Stress* $p_e$ for $q_g$ and $S_c$, then it approves the causal correlation $Gas\_Cont \leftrightarrow Gas\_Prod$.

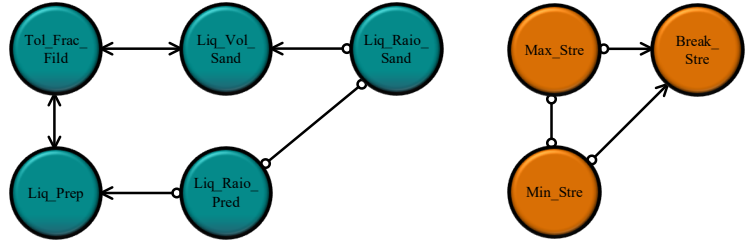

**Fig. 8**. One of causal path in one type of factors: (a) a local causal graph in engineering factors; (b) a local causal graph in geological factors.

**The causal graph existing in the geological/engineering factors.** For the engineering factors, including *Total volume of Fracturing Fluid during Engineering, Liquid Volume of Prepad, Liquid Ratio of Prepad, Liquid Volume of Sand-loaden and Liquid Ratio of Sand-loaden*, we have the following causal graph, shown in Fig. 8(a); For the geological factors, including *Maximum Horizontal Principal Stress, Minimum Horizontal Principal Stress and Breakdown Stress*, we have the causal graph Fig. 8(b). To prove the edge in causal graph in Fig. 8, we apply the tools in Fracturing Engineering to give the proof.

**Proof for Fig. 8(a).** For the Fig. 8(a), the equation of *Total Volume of Proppant during Engineering* $Q_f$ is defined as,

$$Q_f = Q_p + Q_s + Q_r \quad (10)$$

The Eq. (10) can be split as

$$\begin{cases} Q_p = Q_f - Q_r - Q_s \\ Q_s = Q_f - Q_r - Q_p \end{cases} \quad (11)$$

there exists common cause $Q_r$ between $Q_f$, $Q_p$ and $Q_s$, that is, $Liq\_Prep \leftrightarrow Tol\_Frac\_Fild \leftrightarrow Liq\_Sand$.

When the Eq. (10) satisfies $Q_f/Q_f = (Q_p + Q_s + Q_r)/Q_f$, then the relation between $R_p$, $R_s$ and $R_r$ can be represented as,

$$1 = R_p + R_s + R_r \quad (12)$$

According to the Eq. (12), there is no $D$-separation set between $R_p$ and $R_s$, that is, $Liq\_Raio\_Prep$ o— o $Liq\_Raio\_Sand$. And we have for $Q_p$ and $Q_s$

$$\begin{cases} Q_p = R_p Q_f \\ Q_s = R_s Q_f \end{cases} \quad (13)$$

where $R_p$ and $R_s$ are the cause of $Q_p$ and $Q_s$, respectively, that is, $Liq\_Raio\_Prep$ o $\longrightarrow Liq\_Prep$ and $Liq\_Raio\_Sand$ o $\longrightarrow Liq\_Sand$. Then the Fig. 8(a) has been proved by from Eq. (10) to (13).

**Proof for** Fig. 8(b). Moreover, for the Fig. 8(b), the equation of *Breakdown Stress* $p_F$ can be defined with

$$p_F = 3\sigma_y - \sigma_x + \sigma_t + p_s \Longrightarrow p_F = l_1(\sigma_y, \sigma_x, \sigma_t, p_s) \quad (14)$$

Then, we can say the $\sigma_y$ and $\sigma_x$ are the cause of $p_F$, and there is no D-separation between $\sigma_y$ and $\sigma_x$. Then the Fig. 8(b) has been proved by Eq. (14).

**4.4 The Causal Model vs Correlation Model of Predicting Production**

To compare the difference between the production prediction models based solely on correlation and the models based on causality, we select two set of input for the models by correlation and by SHAP with causal explainer.

- **Correlation Variables.** *Gas Content, Gas Saturation, Permeability, Ratio of Critical Desorption Stress to Reservoir Stress* and *Minimum Horizontal Principal Stress*
- **Causal Variables.** *Gas Content, Gas Saturation, Liquid Volume of Prepad, Ratio of Critical Desorption Stress to Reservoir Stress* and *Maximum Horizontal Principal Stress*

For typical machine learning models, Linear regression (LR), SVR, Random Forest (RF), MLP, are employed to predict the CBM production after hydraulic fracturing, with the above two sets of input. The performance of the models can be analyzed with mean squared error (MSE), $R^2$ and mean absolute error (MAE), which are listed in Table 4.

Table 4. Causality Method vs Correlation Method

| | $R^{2\text{train}}$ | MAE$^{\text{train}}$ | MSE$^{\text{train}}$ | $R^{2\text{test}}$ | MAE$^{\text{test}}$ | MSE$^{\text{test}}$ |
|---|---|---|---|---|---|---|
| LR | 0.149 | 0.065 | 0.010 | 0.169 | 0.061 | 0.007 |
| LR$^{\text{causal}}$ | 0.137 | 0.066 | 0.010 | 0.112 | 0.063 | 0.008 |
| SVR | 0.261 | 0.067 | 0.009 | 0.071 | 0.066 | 0.008 |
| SVR$^{\text{causal}}$ | 0.228 | 0.068 | 0.009 | 0.252 | 0.065 | 0.007 |
| MLP | 0.153 | 0.066 | 0.010 | 0.171 | 0.061 | 0.007 |
| MLP$^{\text{causal}}$ | 0.116 | 0.068 | 0.011 | 0.174 | 0.063 | 0.007 |
| RF | **0.864** | **0.027** | **0.001** | 0.082 | 0.065 | 0.008 |
| RF$^{\text{causal}}$ | 0.859 | 0.027 | 0.001 | **0.356** | **0.060** | **0.006** |

**Note**. Method and Method$^{\text{causal}}$ denote machine learning method formalized by correlation and *causality variables, respectively, locating in row indexes. Evaluation$^{train}$ and Evaluation$^{test}$ denote the evaluation index of training process and testing process, respectively, locating in column indexes. And, in which, $\underline{\cdot}$ denote the best value of the corresponding indicator and method.*

As shown in Table 4, the presented models based on causality performs better generalization ability than the models based on correlation analysis. In the training process, the correlation-based models fit the data slightly better than the causality-based models. For example, LR has 8.76% higher training accuracy than LR$^{causal}$; RF has 0.58% higher training accuracy than RF$^{causal}$. On the contrary, in the testing process, the models with causality performs much better that the models with correlation. Here we can see that SVR$^{causal}$, RF$^{causal}$ has improved the prediction accuracy by 18.10% and 27.4%, comparing with SVR and RF respectively. Above all, causality can provide a foundation for the interpretable machining learning to predict the CBM production more accurately.

## 5. Conclusion

Generally, the databased machine learning methods for CBM production prediction are typical black-box models, the performance of which severely depends on the quality and the quantity of the training data. Our method under causality can bring truly relationship to predict production capability of CBM, which provides a novel methodology to interpret the black-box models indirectly. In this article, we design the IICD Algorithm to discover the potential causality from the observed data in a CBM reservoir. The global and the local discussion on the generated causal graph clearly illustrate the causality between geological/engineering factors and production capability. The internal causality in the geological factors and the engineering factors are compared with the actual mechanism of Fracturing Engineering, the result shows that the discovered causality is in consistent with our cognition on CBM development.

Moreover, the calculation results show that the machine learning models under causality perform better accuracy that the traditional models based on correlation analysis. Then we provide a foundation for interpretable machine learning to capture the causality between the factors and production capability of CBM after fracturing. The presented method in this article can help improve the design of machine learning models to predict the production, which is significant in the case that we cannot obtain enough high-quality training data.

## Acknowledgements

This paper is supported by the International Cooperation Program of Chengdu City (No. 2020-GH02-00023-HZ) and the Innovation Seedling Project of Sichuan Province "Research on prediction method of oil and gas field development index based on interpretable machine learning"(NO:2022034)